%% file: main.tex
\pdfoutput=1
\documentclass[11pt]{article}

\usepackage[]{acl}

\usepackage{times}
\usepackage{latexsym}

\usepackage[T1]{fontenc}
\usepackage[utf8]{inputenc}
\usepackage{microtype}
\usepackage{inconsolata}

\usepackage{adjustbox}
\usepackage{booktabs}
\usepackage{makecell}
\usepackage{multirow}
\usepackage{amsmath}
\usepackage{comment}
\usepackage{caption}
\usepackage{subcaption}
\usepackage{cleveref}

\title{
    Formulation Comparison for Timeline Construction using LLMs
}

\author{
    Kimihiro Hasegawa \hspace{0.5cm} {\bf Nikhil Kandukuri} \\ {\bf Susan Holm} \hspace{0.5cm} {\bf Yukari Yamakawa} \hspace{0.5cm} {\bf Teruko Mitamura} \\
    Language Technologies Institute, Carnegie Mellon University \\
    \texttt{\{kimihiro,teruko\}@cs.cmu.edu, \{nkanduku,sh4s,yukariy\}@andrew.cmu.edu}
}

\begin{document}
\maketitle
\begin{abstract}

Constructing a timeline requires identifying the chronological order of events in an article. 
In prior timeline construction datasets, temporal orders are typically annotated by either event-to-time anchoring or event-to-event pairwise ordering, both of which suffer from missing temporal information. 
To mitigate the issue, we develop a new evaluation dataset, TimeSET, consisting of single-document timelines with document-level order annotation.
TimeSET features saliency-based event selection and partial ordering, which enable a practical annotation workload. 
Aiming to build better automatic timeline construction systems, we propose a novel evaluation framework to compare multiple task formulations with TimeSET by prompting open LLMs, i.e., Llama 2 and Flan-T5.
Considering that identifying temporal orders of events is a core subtask in timeline construction, we further benchmark open LLMs on existing event temporal ordering datasets to gain a robust understanding of their capabilities.
Our experiments show that (1) NLI formulation with Flan-T5 demonstrates a strong performance among others, while (2) timeline construction and event temporal ordering are still challenging tasks for few-shot LLMs.\footnote{Our code and data are available under~\url{https://github.com/kimihiroh/timeset}.}
\end{abstract}

\input{sections/intro}
\input{sections/related-work}
\input{sections/annotation}
\input{sections/formulation-comparison}
\input{sections/benchmark}
\input{sections/conclusion}
\input{sections/limitation-and-ethical-consideration}


\bibliography{anthology,custom}

\clearpage
\appendix

\input{sections/appendix}

\end{document}

%% file: sections/intro.tex
\section{Introduction}
\label{sec:intro}

\input{figures/overview_formulation_comparison}

Timeline construction requires structuring events in chronological order based on textual input.
This task has implications for practical applications, such as assisting well-informed decisions~\citep{minard-etal-2015-semeval}, text analysis~\citep{toro-isaza-etal-2023-fairy}, or script induction~\citep{sancheti-rudinger-2022-large}.
In a news article, events are not always described in chronological order~\citep{zwaan1995dimensions}.
For instance, from~\autoref{fig:overview}:
\begin{quote}
    \textit{Amir Khan was given a ... ban for doping}
\end{quote}
\noindent ``\textit{ban}'' is mentioned before ``\textit{doping}'', but the temporal order is ``\textit{doping}'' $\rightarrow$ ``\textit{ban}.''

Prior work has addressed timeline construction as a time anchoring task: identify the corresponding time expression of each event as its \textit{anchor} and compare the anchors to order events~\citep{minard-etal-2016-meantime}.
However, sequential events can be anchored to the same coarse time range~\citep{naik-etal-2019-tddiscourse}.
Also from~\autoref{fig:overview}:
\begin{quote}
    \textit{testing positive for ostarine ... in a urine sample following his ... loss ... in February}
\end{quote}
\noindent both ``\textit{testing}'' and ``\textit{loss}'' are anchored to ``\textit{February},'' which fails to capture the temporal order, ``\textit{loss}'' $\rightarrow$ ``\textit{testing}.''
Handling such cases necessitates the comprehension of context.

In parallel, event temporal ordering has been actively explored as a context-based task without exclusive attention to time expressions. 
Conventionally, event temporal ordering is formulated as a pairwise classification task. 
While this approach could cover every pair, the annotation cost for $\binom n2$ pairs gets expensive as the number of events $n$ increases, contrary to the time anchoring approach~\citep{reimers-etal-2016-temporal}.
In practice, previous work focuses only on local pairs (e.g., within adjacent sentences) to reduce the cost~\citep{ning-etal-2018-multi}, which makes the datasets subopitmal to evaluate timeline construction systems.

In terms of task design, other task formulations, such as Natural Langauge Inference (NLI) by TemporalNLI~\citep{vashishtha-etal-2020-temporal} or Machine Reading Comprehension (MRC) by TORQUE~\citep{ning-etal-2020-torque}, have also been explored to assess models' capabilities on event temporal reasoning.

In principle, models can only be compared within one dataset based on its task formulation, while a model comparison across datasets is not supported.
Considering that multiple datasets assess the same event temporal ordering capability, one question arises: Which task formulation best elicits the capabilities of models?
This question gains further importance with the surge of LLMs, which can be evaluated on multiple datasets in a few-shot setting.
More specifically, as shown in~\autoref{fig:overview}, LLMs can be prompted to identify one pairwise relation at a time (e.g., ``\textit{Relation between ban and retired is}'',  ``\textit{AFTER}'') or multiple relations in one question-answer pair (e.g., ``\textit{What happened after testing?}'' ``\textit{retired, ban}'').

In this paper, we first propose a new evaluation dataset, \textbf{TimeSET} (\textit{\textbf{Time}line for \textbf{S}alient \textbf{E}ven\textbf{T}s}), to address the lack of testbed for context-based timeline construction.
TimeSET consists of diverse topics of Wikinews\footnote{\url{https://www.wikinews.org/}} articles with single-document timeline annotation based on document-level pairwise orders.
We introduce two features in TimeSET, which enables a practical annotation workload: saliency-based event selection, motivated by the similarity between Information Extraction (IE) and Summarization~\citep{grishman1999cross}, and partial-order annotation, inspired by script annotation~\citep{sakaguchi-etal-2021-proscript-partially} (Section~\ref{sec:annotation}).

Secondly, we propose a novel evaluation framework for timeline construction that enables comparison across models and task formulations.
Specifically, we cast TimeSET into four task formulations, Natural Language Inference (NLI), Pairwise, Machine Reading Comprehension (MRC), and Timeline (illustrated in~\autoref{fig:overview}).
For the same document-timeline pair, each formulation prompts LLMs in its manner a different number of times to identify the temporal orders of a timeline from a document.
Intuitively, LLMs favor one formulation over others depending on its proximity to the exposed tasks during pretraining or tuning.
With this framework, we address our first research question, \textbf{(1) ``Which task formulation better elicits LLMs' capabilities for timeline construction?,''} especially targeting open LLMs (i.e., Llama 2~\citep{touvron2023llama} and Flan-T5~\citep{chung2022scaling}) (Section~\ref{sec:formulation_comparison}).

Furthermore, we address our second research question, \textbf{(2) ``How good are open LLMs at identifying the order of events?,''} by benchmarking open LLMs on existing datasets, namely, TemporalNLI, MATRES~\citep{ning-etal-2018-multi}, TDDiscourse~\citep{naik-etal-2019-tddiscourse}, and TORQUE.
Since event temporal ordering is a core subtask of timeline construction, benchmarking open LLMs in in-context learning~\citep{brown2020language} and comparing them with smaller-sized fine-tuned models help to gain robust insights for further development of timeline construction systems (Section~\ref{sec:benchmark}). 

In summary, our contributions are three-fold:
\begin{itemize}
    \item We develop a new evaluation dataset, TimeSET, to support context-based timeline construction.
    The dataset is available under an open license.
    \item Addressing the research question \textbf{(1)}, we propose a novel evaluation framework that enables cross-model and cross-formulation comparisons. Our experiment using TimeSET indicates that NLI formulation with Flan-T5 demonstrates a strong performance among others.
    \item For the research question \textbf{(2)}, we benchmark open LLMs in a few-shot setting on existing event temporal ordering datasets. We find that few-shot LLMs underperform smaller-sized fine-tuned models.
\end{itemize}

%% file: figures/overview_formulation_comparison.tex
\begin{figure*}[t]
    \centering
    \includegraphics[width=\textwidth]{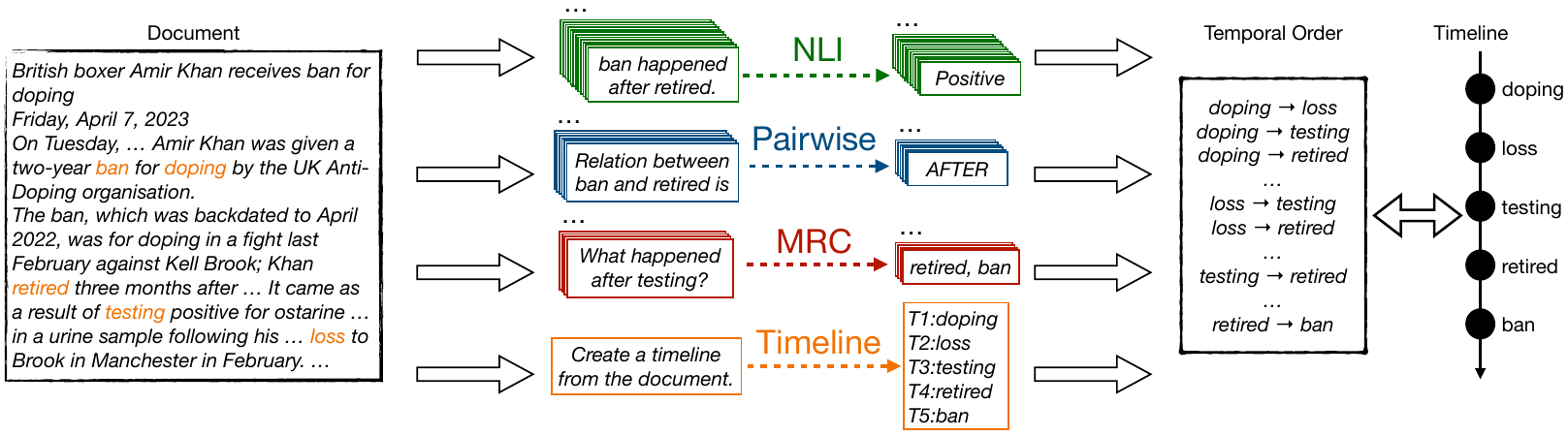}
    \caption{A comparison overview of the four formulations we study in this paper: NLI, Pairwise, MRC, and Timeline. A document with event annotation is converted into a different number of prompts following each formulation, and the predictions are interpreted as pairwise temporal orders in each formulation's manner to form a timeline.}
    \label{fig:overview}
\end{figure*}


%% file: sections/related-work.tex
\section{Related Work}
\label{sec:related_work}

In this section, we show previous studies in connection with our new dataset, formulation comparison, and benchmarking.

\subsection{Temporal Order Annotation}
\label{subsec:dataset_comparison}

In the area of event-centric information extraction~\citep{chen-etal-2021-event}, temporal orders have been annotated by mainly two approaches: time anchoring and pairwise ordering. 
Time-anchoring annotation has been used in developing timeline construction datasets, such as TimeLine shared task~\citep{minard-etal-2015-semeval} and MEANTIME~\citep{minard-etal-2016-meantime}.
Several studies on event temporal ordering also employ this approach: Event StoryLine Corpus~\citep{caselli-vossen-2017-event}, ~\citet{cheng-miyao-2018-inducing}, MAVEN-ERE~\citep{wang-etal-2022-maven}, and TIMELINE~\citep{alsayyahi-batista-navarro-2023-timeline}.
Connecting each event with a time expression requires only a linear-time annotation~\citep{reimers-etal-2016-temporal}, although it potentially misses some temporal orders that can be identified from context information.

Pairwise-ordering annotation, which adds temporal relations to each pair of events, has supported a variety of event temporal ordering datasets: TimeBank~\citep{pustejovsky2003timebank}, TBDense~\citep{cassidy-etal-2014-annotation}, and MATRES, \textit{inter alia}.
At the expense of document-level information, they typically limit the annotation scope within a few sentence windows to address the annotation sparsity and alleviate the expensive cost of quadratic-pair annotation.
While TDDiscourse augments long-distance relation annotation on top of TBDense, the additional annotation covers only a subset of all pairs, which prevents the use of evaluating timeline construction systems.
Also, pairwise annotation datasets usually employ an inclusive definition of event\footnote{a cover term for situations that happen or occur ... can be punctual or last for a period of time ... those predicates describing states
or circumstances in which something obtains or holds true}~\citep{pustejovsky2003timeml} that does not fit into a summary-oriented timeline. 

Contrary to these previous studies, our TimeSET targets timeline construction with document-level pairwise annotation, featured by the saliency-based event selection and partial-order annotation.
Temporal dependency graph~\citep{yao-etal-2020-annotating} shares a similar goal with our work to balance between cost efficiency and document-level annotation using dependency-based annotation among events and time expressions.
As it aims to accurately represent temporal information, it may contain details that summary-like timelines may not need.

\subsection{Formulation Comparison}
Lately, there has been a trend of developing a unified task formulation for various NLP tasks, such as QA~\citep{khashabi-etal-2020-unifiedqa, rogers2023qa}, NLI~\citep{poliak-etal-2018-collecting}, text-to-text generation~\citep{raffel2020exploring}, and prompting~\citep{liu2023pre}. 
A few works have explored the meta-comparisons across task formulations in IE:
\citet{cohen2022supervised} compared the span-prediction and classification objectives for entity-centric relation classification tasks in a supervised setting.

Our work focuses on the timeline construction task with LLMs in an in-context learning setting, where we recast the task into different compositionally~\citep{dziri2023faith} of temporal ordering tasks.

\subsection{LLM for IE}
LLMs have been evaluated on benchmark datasets, such as MMLU~\citep{hendrycks2021measuring}, HELM~\citep{liang2022holistic}, and BBH~\citep{suzgun-etal-2023-challenging}, to name a few. 
Among IE tasks, the primary focus is paid on entity-centric IE tasks~\citep{xu-etal-2023-unleash, wadhwa-etal-2023-revisiting}, along with some studies on general event-centric IE tasks~\citep{ma-etal-2023-shot, ma-etal-2023-large}.

Specifically for temporal understanding, existing works mostly focus on OpenAI's proprietary models~\citep{ouyang2022training, chatgpt2022, gpt42023}: temporal ordering~\citep{chan2023chatgpt, yuan-etal-2023-zero}, NLI-style temporal reasoning~\citep{feng-etal-2023-generic}, or temporal grounding~\citep{qiu2023large}. 
A few works that evaluate open LLMs investigate other types of temporal-related tasks than event temporal ordering:
\citet{tao2023eveval} evaluated BLOOM models~\citep{workshop2022bloom} and Flan-T5 on a suite of event semantic processing tasks.
\citet{qiu2023large} investigated Llama families on temporal grounding tasks, using MCTACO~\citep{zhou-etal-2019-going}, CATERS~\citep{mostafazadeh-etal-2016-caters}, and QA TempEval~\citep{llorens-etal-2015-semeval}.

In contrast, our benchmark experiment focuses specifically on the event temporal ordering task with open LLMs and compares them with smaller-sized fine-tuned models.
Our findings can provide insights complementary to the prior studies to develop better timeline construction systems.

%% file: sections/annotation.tex
\section{TimeSET}
\label{sec:annotation}

We propose a new evaluation dataset for timeline construction, TimeSET, consisting of diverse topics from Wikinews articles.
To make TimeSET suitable for the timeline construction task with a tractable annotation workload, we introduce two features: saliency-based event selection and partial-ordering annotation.

\paragraph{Salient Event:} This design choice is motivated by multiple prior studies. First,~\citet{minard-etal-2015-semeval} points out that not all events in a news article are worth being mentioned for a timeline, where the importance of focusing on \textit{core} events are also discussed in~\citet{chambers-2017-behind}.
Second, we focus on summary-based salient events~\citep{liu-etal-2018-automatic}, based on the view that ``\textit{IE and Summarization are two sides of a coin ... different emphasis of output}''~\citep{grishman1999cross}.\footnote{The summarization-based approach also brings a practical benefit of reducing target events for order annotation.}

\paragraph{Partial Ordering:} To reduce the annotation demand, we employ a partial-ordering approach, inspired by proScript~\citep{sakaguchi-etal-2021-proscript-partially}.
Specifically, each event is connected with one preceding and one following event, except for the oldest and newest events in each timeline.
\citet{yao-etal-2020-annotating} takes a similar strategy of annotating temporal dependencies among events and time expressions, whereas we focus exclusively on salient events. 

Additionally, we annotate arguments to provide the potential to disambiguate events. 
We refer to Propbank~\citep{palmer-etal-2005-proposition} for role names.
 
\subsection{Annotation Guideline}
\label{sec:annotation_guideline}
We annotated events and temporal links, as well as arguments.
We used brat~\citep{stenetorp-etal-2012-brat} as our main annotation interface (Figure~\ref{fig:brat}).

\textbf{Event:} We define events as actions, happenings, and occurrences, excluding states, reporting events, grammatical events~\citep{minard-etal-2015-semeval}, and events with a modality/realis other than actual~\citep{tonelli2014newsreader}. 
We set two criteria for saliency: (1) if the event can appear in a short summary and (2) if the event is relevant to the document's title. 
When multiple coreferring mentions exist within one document, annotators chose the most representative one.

\textbf{Temporal Link:} We added pairwise links to represent chronological order between pairs of annotated events (\textit{AFTER} link).
We focused on the start time of events as~\citet{ning-etal-2018-multi}. 
In addition to \textit{AFTER} link, we introduced a coexistent relation (\textit{COEX} link) to represent a fuzzy relation where two events happened around the same time but their chronological order is not mentioned explicitly in the context. 
In addition to brat, we used our new visualization tool to ensure event connectivity (Figure~\ref{fig:vis}). 

\subsection{Document Collection}
We chose news articles as our target document type because they commonly discuss a sequence of events. 
As our source, we selected Wikinews considering its open license (CC BY 2.5) for easy redistribution.
We collected 50 English articles considering the following criteria. 
(1) Length: Referring to~\citet{yao-etal-2020-annotating}, we collected documents with mainly more than 300 words.
(2) Document Creation Time: To investigate possible data contamination~\citep{sainz-etal-2023-nlp}, we set September 2022\footnote{This boundary was decided based on the data cutoff date for the pretraining of Llama 2.} as the boundary for categorizing documents into ``new'' and ``old.''
(3) Category: To ensure diversity, we collected documents from 27 topics (\ref{appx:annotation_topics}), which is 5 times more than prior work~\citep{minard-etal-2015-semeval}.

\subsection{Annotation Procedure and Statistics}

\input{tables/dataset_stats_simple} 
\input{tables/formulation_examples} 

We developed our annotation guideline by annotating small samples of articles with two expert annotators. 
After a few rounds of annotation and discussion with a small set of 10 documents, the annotation guideline was finalized and the experts used it to annotate 40 articles in total, with 3 overlaps to calculate agreement.
The initial 10 adjudicated documents used for guideline development are treated as the development set and the remaining 40 documents as the test set.
The statistics are available in~\autoref{table:dataset_stats_simple}.

From the overlapping documents, we calculate inter-annotator agreements. 
For event spans, the Dice coefficient (pairwise F1)~\citep{minard-etal-2015-semeval} is $0.74$ after resolving coreference. 
For temporal relations, after coreference resolution, the temporal awareness score~\citep{uzzaman-etal-2013-semeval} (See \ref{appx:annotation_analysis}) is $0.50$ for all events and $0.90$ for common events.~\footnote{Common events are the ones both experts annotated as events, which accounts for $75\%$ of all events.}
The inevitable subjectivity in saliency affects the IAA of event spans, which is propagated to the score of all-event temporal relations. 
However, the high score of common-event temporal relations indicates the high quality of our temporal relation annotation. 
Further discussion is in Appendix~\ref{appx:annotation_analysis}.

%% file: tables/dataset_stats_simple.tex
\begin{table}[!t]
\centering
\begin{adjustbox}{width=0.45\textwidth}
\begin{tabular}{ c c c c}
    \toprule
    Docs (Words) & Events & Relations & Args\\
    \midrule
    50 (437.9) & 356 (7.1) & 314 (6.3) & 654 (13.1) \\
    \bottomrule
\end{tabular}
\end{adjustbox}
\caption{TimeSET statistics. The number of documents (avg. words), events (avg), relations (avg), and arguments (avg). Averages are per document.}
\label{table:dataset_stats_simple}
\end{table}

%% file: tables/formulation_examples.tex
\begin{table*}[!t]
\centering
\begin{adjustbox}{width=0.95\textwidth}
\begin{tabular}{c l l c }
    \toprule
    Formulation & Prompt & Raw Prediction & Predicted Order(s) \\
    \midrule 
    NLI & \textit{[context] bought happened after went. Is this true?} & \textit{Yes} & \textit{went} $\rightarrow$ \textit{bought} \\
    Pairwise & \textit{[context] Relation between went and bought is} & \textit{BEFORE} & \textit{went} $\rightarrow$ \textit{bought} \\
    MRC & \textit{[context] What happened after visited?} & \textit{went, bought} & \makecell{\textit{visited} $\rightarrow$ \textit{went},\\ \textit{visited} $\rightarrow$ \textit{bought}} \\
    Timeline & \textit{[timeline] Create a timeline.} & \textit{T1: visited \textbackslash n T2: went \textbackslash n T3: bought} & \textit{visited} $\rightarrow$ \textit{went} $\rightarrow$ \textit{bought} \\
    \bottomrule
\end{tabular}
\end{adjustbox}
\caption{An example prompt and expected output with its corresponding order(s) for each formulation. Here, the context is ``\textit{Yesterday, I \textbf{went} to a bookstore and \textbf{bought} a novel. The last time I \textbf{visited} there was a month ago.}'' (\textbf{Events} are in boldface.)} 
\label{table:formulation_examples}
\end{table*}

%% file: sections/formulation-comparison.tex
\section{Formulation Comparison}
\label{sec:formulation_comparison}

\input{figures/result_formulation_comparison_base}

Addressing ``Which task formulation better elicits LLMs’ capabilities for timeline construction?,'' we propose a novel evaluation framework to identify a better combination of a model and a formulation for timeline construction.
As illustrated in~\autoref{fig:overview}, we cast TimeSET in four formulations, namely, NLI, Pairwise, MRC, and Timeline. 
Each formulation decomposes the timeline construction task to different compositional degrees of temporal ordering tasks.
This comparison is motivated by our intuition that LLMs prefer one formulation to others due to the similarity between formulations and training data.

\subsection{Formulation}
In each formulation, an input consists of a context with event annotation and formulation-specific prompt in a prompt template, which is fed to LLMs to make a different form of prediction.\footnote{As event annotation is given in input, our experiment is based on our high-quality temporal order annotation.}
\autoref{table:formulation_examples} shows an example input-output pair for each formulation.
(Actual examples in Appendix~\ref{appx:prompt_template}).

\paragraph{NLI}: 
An input consists of a context (premise) and a statement with a pair of events and a relation (hypothesis). 
The output is a binary label indicating the truth value of the statement. 
We generate permutations for each pair of events and each relation, resulting in $_nP_2 \times k = n \times(n-1)\times k$ instances from each document. 
($n$ and $k$ represent the number of events in a document and the number of temporal relations types, respectively.\footnote{In this experiment, $k=3$: AFTER, BEFORE, and COEX})

\paragraph{Pairwise}: 
An input consists of a context and a pair of events, and its output represents the relation that holds between the two events. 
We generate permutations from each pair, resulting in $_nP_2 = n \times (n-1)$ instances from each document. 

\paragraph{MRC}: 
A context and a question about an event and a relation form an input. 
The output consists of a list of event(s), where the order of events in the output does not imply temporal orders.
We generate $n \times k$ instances from each document. 

\paragraph{Timeline}: 
An input is essentially a context, and its output is a list of events, where the order is interpreted as the chronological order. 
Each document forms one instance. 
We use each layer in breadth-first search traversal as one time range in a timeline. 

\subsection{Model}
We target two families of LLMs, Llama 2 and Flan-T5, to cover model types as diverse as possible under an academic computational environment so that our findings can be generalized to other LLMs.
The following are the criteria: 
(1) Transparency: We narrow our targets to open models (e.g., Flan-T5).~\footnote{We exclude closed models, such as GPT series~\citep{chatgpt2022, gpt42023}, due to their difficulties in reproducing results and investigating what characteristics contribute to performance.}
(2) Architecture: We choose two major architectures: decoder-only models (e.g., Llama 2) and encoder-decoder models (e.g., Flan-T5). 
(3) Model Variant: We prioritize model families that provide different-sized and variously-tuned models. 
For Llama 2, we test Llama 2 Chat~\citep{touvron2023llama}, which is an RLHF-tuned model of Llama 2, and CodeLlama~\citep{roziere2023code}, which reportedly performs well in structured prediction tasks~\citep{wang-etal-2023-code4struct}. 
For Flan-T5, we pick T5~\citep{raffel2020exploring}, which is the original model before Flan-T5's supervised instruction-tuning.
We test these models in an in-context learning setting.

\subsection{Prompt Template}
With in-context learning, prompts have a large impact on performance~\citep{gonen2022demystifying}. 
To account for the effect of prompts, we manually curated 10 templates for each task, referring to P3~\citep{sanh2022multitask}, the Flan Collection~\citep{longpre2023flan}, and others~\citep{madaan-etal-2022-language, zhang-etal-2023-causal} (Examples in Appendix~\ref{appx:prompt_template}). 
We select demonstrations randomly from each development split.
The number of demonstrations is 0, 1, or 2, constrained by the models' maximum input length. 
In evaluation, we report median performance across prompt templates and different numbers of demonstrations, following~\citet{perez2021true, sanh2022multitask}.  
Other experimental details are in Appendix~\ref{appx:formulation_experiment_detail}.

\subsection{Results and Analysis}
\input{figures/analysis_summary}
We use a document-wise pairwise F1 score as our metric.
\autoref{fig:result_formulation_comparison_base} shows the overall result with medians with their interquartile range (Q3 - Q1). 
Among others, the combination of NLI formulation with Flan-T5 shows a strong performance, while more compositional formulations (MRC and Timeline) perform better on average across models.
However, no single formulation outperforms others for all models. 
To put it differently, the same model performs differently under different task formulations. 
Even models from the same model family produce different results for each formulation.
This implies that one model should be evaluated on multiple formulations to elicit its capability or multiple models should be employed for assessing one formulation. 
In terms of computational cost, the Timeline formulation with Llama 2 Chat would be a practical choice, since it requires only one prediction for each document, while Flan-T5 in the NLI formulation requires $k \times n \times (n-1)$ times of predictions.

Checking the Llama 2 family, the best performances are on par across formulations.
Llama 2 Chat (7B) generally improves the median performance across formulations from Llama 2 (7B).
CodeLlama (7B) produces relatively better results in more structured formulations.

Looking at the Flan-T5 family, on one hand, Flan-T5 (3B) shows strong results in NLI formulation and better results in the pairwise formulation than T5 (3B). 
On the other hand, Flan-T5 (3B) underperforms T5 (3B) in the MRC and Timeline formulations. 
It could be attributed to Flan-T5's instruction-tuning data, where various types of NLI datasets are included.
However, since the data also contains diverse MRC datasets, it may not be the main reason. 
More detailed analysis, for instance, in terms of underlying reasoning skills, would be required, and we leave it to future work.

Furthermore, we study the effects of model size and document characteristics (See more analysis in Appendix~\ref{appx:formulation_comparison_analysis}).

\paragraph{Do Larger Models Perform Better?}
Figure~\ref{fig:analysis_size} shows the median performance of difference sizes of Llama 2 and Flan-T5. The result indicates that larger models perform better regardless of formulations in general. This tendency is more evident for Flan-T5. 

\paragraph{Do Models Have Preference on Documents?}
We further investigate the breakdown performance based on document characteristics, namely, Document creation time (DCT) and document length.
\autoref{fig:analysis_date} and~\autoref{fig:analysis_num_word} shows the results.
Llama-2 (7B) performs slightly better on documents written before its pretraining data cut-off date, but the difference is marginal compared to the variance across prompt templates. 
Performance gets worse with longer documents.
This trend is more pronounced with formulations that identify multiple orders at once, like the Timeline formulation.

Overall, timeline construction is still a challenging task for few-shot LLMs, and we hope our evaluation framework with TimeSET can facilitate future research.

%% file: figures/result_formulation_comparison_base.tex
\begin{figure*}[t]
    \centering
    \includegraphics[width=\textwidth]{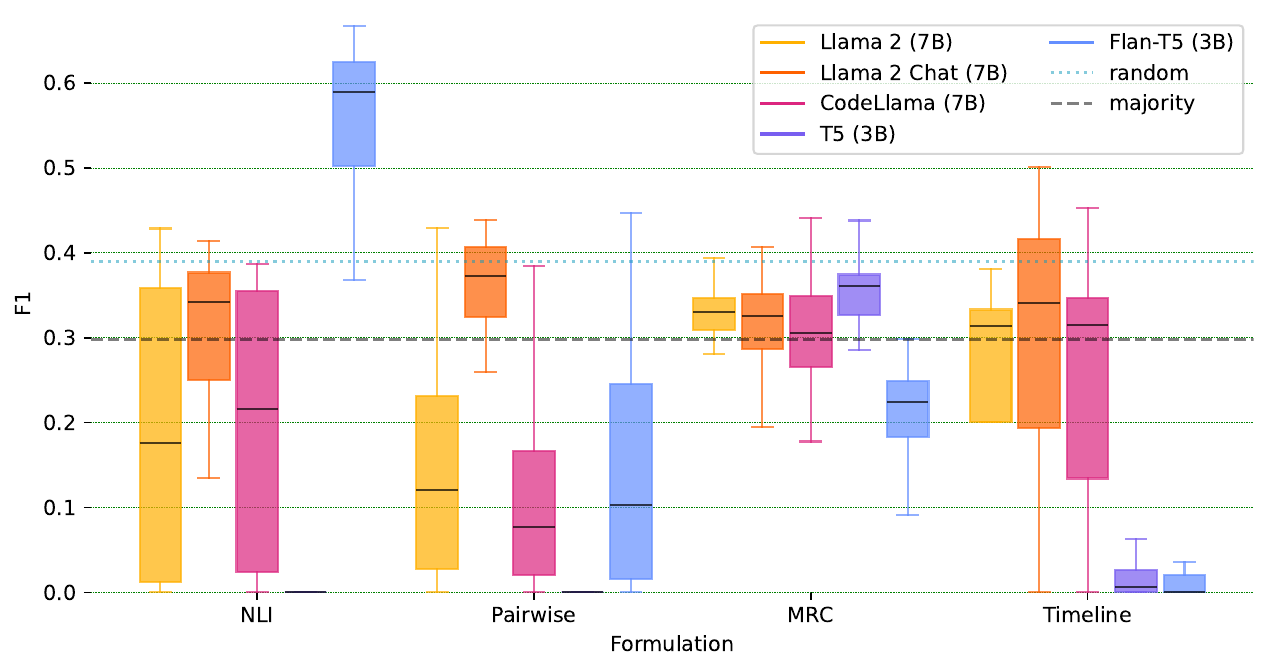}
    \caption{Formulation comparison result with TimeSET. In each boxplot, one data point represents a combination of a prompt template and the number of demonstrations.}
    \label{fig:result_formulation_comparison_base}
\end{figure*}

%% file: figures/analysis_summary.tex
\begin{figure*}[t]
    \centering
    \begin{subfigure}[b]{0.329\textwidth}
        \centering
        \includegraphics[width=\textwidth]{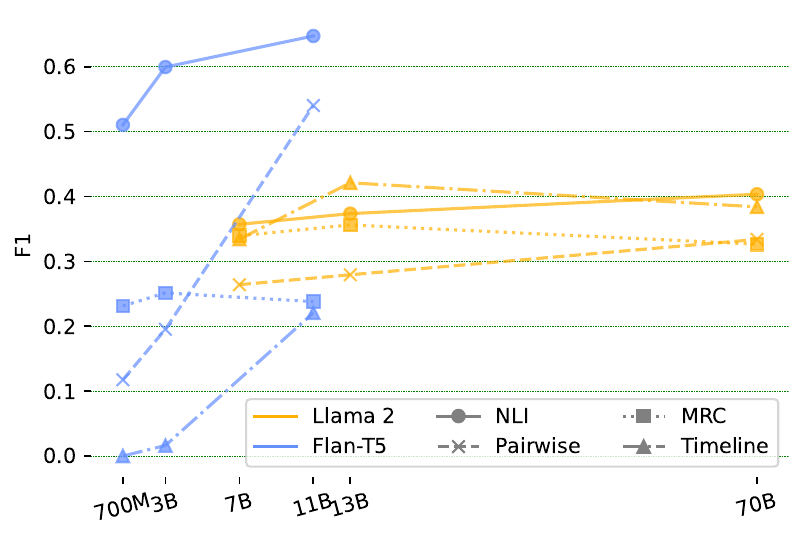}
        \caption{Model size}
        \label{fig:analysis_size}
    \end{subfigure}
    \hfill
    \begin{subfigure}[b]{0.329\textwidth}
        \centering
        \includegraphics[width=\textwidth]{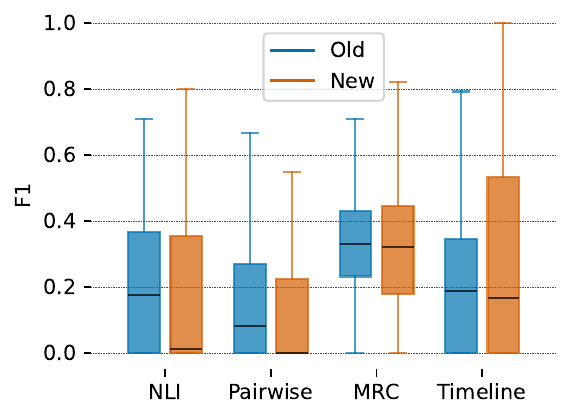}
        \caption{DCT with Llama 2 (7B)}
        \label{fig:analysis_date}
    \end{subfigure}
    \hfill
    \begin{subfigure}[b]{0.329\textwidth}
        \centering
        \includegraphics[width=\textwidth]{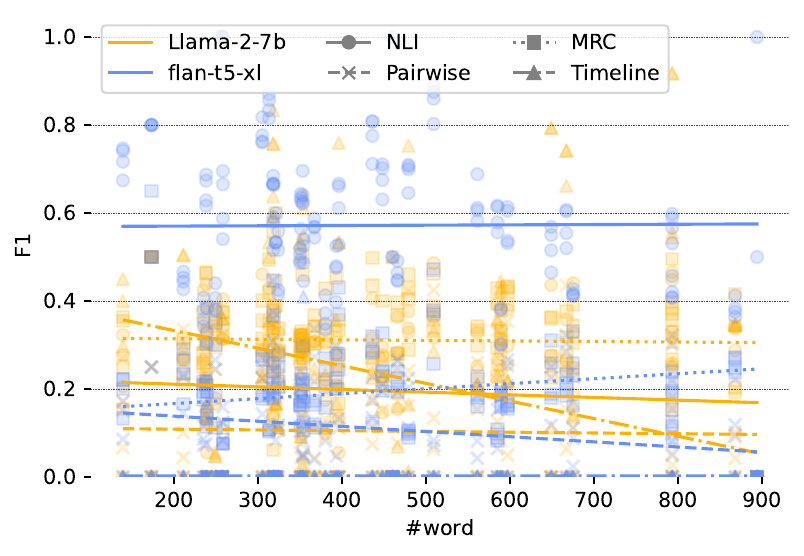}
        \caption{\#word}
        \label{fig:analysis_num_word}
    \end{subfigure}
    \caption{Analysis Overview}
    \label{fig:analysis_overview}
    
\end{figure*}

%% file: sections/benchmark.tex
\section{Benchmarking on Existing Datasets}
\label{sec:benchmark}
\input{tables/result_benchmark}
To better understand LLMs' capabilities in the core subtask of timeline construction, we benchmark LLMs on existing event temporal ordering datasets.

\subsection{Dataset}
Among a variety of event temporal ordering datasets, we chose four datasets based on their diverse characteristics, especially in terms of their formulations. Following are brief descriptions of the datasets:
TemporalNLI consists of automatically-recast pairwise datasets. Among the subsets, we use the recast version of TempEval3~\citep{uzzaman-etal-2013-semeval}.
MATRES is one of the standard datasets with context-based pairwise annotation within adjacent sentences. 
TDDiscourse is also a pairwise temporal ordering dataset. We include this dataset as it contains document-level annotation. We use the manually-annotated set, TDDMan, in our experiment.
TORQUE consists of a machine reading comprehension-style dataset for event temporal reasoning.
Dataset splits and examples are provided in Appendix~\ref{appx:benchmark_dataset}.

\subsection{Model and Tuning Method}
Our target LLMs, prompt template, and fine-tuned models are as follows.

\paragraph{LLMs:}
Similar to our formulation comparison experiments, we evaluate two families of models in a few-shot setting: Llama 2 and T5, along with their variants, Llama 2 Chat and Flan-T5.

\paragraph{Prompt Template:}
We use the corresponding set of prompt templates in the formulation comparison experiments. 
Specifically, NLI for TemporalNLI, Pairwise for MATRES and TDDiscourse, and MRC for TORQUE.

\paragraph{Finetuned Models:}
Three aspects are considered in fine-tuned models: 
(1) Architecture: We compare an encoder-only model (DeBERTaV3,~\citet{he2023debertav}), a decoder-only model (Llama 2), and an encoder-decoder model (Flan-T5). 
(2) Learning objective: Language modeling loss is used for models with decoders and either classification loss or sequence tagging loss is applied to encoder-only models, depending on the datasets (More details in Appendix~\ref{appx:benchmark_model_arch}).
(3) Tuning method: To update parameters, we employ full-parameter fine-tuning (FT) and parameter-efficient fine-tuning (PEFT), specifically LoRA~\citep{hu2022lora}.
We also include the SoTA models~\citep{mathur-etal-2022-doctime, zhou-etal-2022-rsgt, man2022selecting, huang-etal-2022-understand}. 
Further experimental details are in Appendix~\ref{appx:benchmark_experiment_detail}.

\subsection{Results}
\autoref{table:result_benchmark} summarizes the results.\footnote{We could not get the result of the Llama 2 (7B) model for the TDDiscourse dataset because of an out-of-memory issue with our one GPU setting (\ref{appx:benchmark_experiment_detail}).}
In almost all cases, fine-tuned methods (FT, PEFT, and Existing Works) outperform LLMs with few-shot learning (ICL) despite the difference in model size. 
This finding aligns with previous studies about general event-centric IE tasks~\citep{ma-etal-2023-large, li2023evaluating}, where it is still challenging for LLMs to perform such tasks with only a few demonstrations. 
In both ICL and PEFT settings, larger models perform generally better than smaller models in each model family, although the differences are smaller than the performance differences between ICL and FT/PEFT. 
The performances of PEFT models are comparable with, or sometimes even better than, those of full parameter-tuned models, which shows the usefulness of PEFT for event temporal ordering tasks.
The difference between model architectures/learning objectives has a marginal effect on performance, although DeBERTaV3 demonstrates its efficacy considering its relatively smaller size.
We use the same hyperparameter search space across all fine-tuned models, but individual hyperparameter searches for each model could possibly boost their performance. 
The best hyperparameters in our experiments are in~\autoref{table:hyperparameters}.
Overall, temporal ordering tasks are still challenging for few-shot LLMs, and hopefully, our results help the research community to gain understanding on their capabilities.

%% file: tables/result_benchmark.tex
\begin{table*}[!t]
\centering
\begin{adjustbox}{width=0.95\textwidth}
\begin{tabular}{c l c c c c}
    \toprule
    \makecell[c]{Tuning\\Method} & Model (\#parameter) & \makecell[c]{TemporalNLI\\(Accuracy)} & \makecell[c]{MATRES\\(micro-F1)} & \makecell[c]{TDDiscourse\\(micro-F1)} & \makecell[c]{TORQUE\\(Exact Match)} \\
    \midrule
    \multirow{2}{*}{FT} & DeBERTaV3 (440M) & 0.531 & 0.736 & 0.439 & 0.493 \\
     & Flan-T5 (770M) & 0.524 & 0.744 & 0.234 & 0.407 \\
    \midrule
    \multirow{4}{*}{PEFT} & DeBERTaV3 (440M) & 0.211 & 0.743 & 0.403 & 0.510 \\
    & Flan-T5 (770M) & 0.550 & 0.763 & 0.243 & 0.463 \\
    & Flan-T5 (3B) & 0.550 & 0.750 & 0.437 & 0.509 \\
    & Llama-2 (7B) & 0.539 & 0.717 & --- & 0.436 \\
    \midrule
    \multirow{7}{*}{ICL} & Llama 2 (7B) & 0.269 & 0.139 & 0.147 & 0.118 \\
    & Llama 2 (13B) & 0.336 & 0.457 & 0.204 & 0.086 \\
    & Llama 2 (70B) & 0.329 & 0.290 & 0.033 & 0.158 \\
    & Llama 2 Chat (7B) & 0.340 & 0.473 & 0.214 & 0.036 \\
    & Flan-T5 (3B) & 0.337 & 0.311 & 0.063 & 0.028 \\
    & Flan-T5 (11B) & 0.375 & 0.386 & 0.124 & 0.034 \\
    & T5 (3B) & 0.0 & 0.0 &  0.0 & 0.0\\
    \midrule
    \multicolumn{2}{c}{Existing Works} & \makecell[c]{0.625\\\citep{mathur-etal-2022-doctime}} & \makecell[c]{0.840\\\citep{zhou-etal-2022-rsgt}} & \makecell[c]{0.511\\\citep{man2022selecting}} & \makecell[c]{0.522\\\citep{huang-etal-2022-understand}} \\
    \bottomrule
\end{tabular}
\end{adjustbox}
\caption{Benchmarking experiment results. FT: full-parameter finetuning. PEFT: parameter-efficient fine-tuning (LoRA). ICL: in-context learning (zero/few-shot learning).}
\label{table:result_benchmark}
\end{table*}


%% file: sections/conclusion.tex
\section{Conclusion}
Toward the end goal of timeline construction, we develop a new evaluation dataset, \textbf{TimeSET}.
It consists of single-document timelines with context-based annotation, characterized by saliency-based event selection and partial-order annotation.
Using TimeSET, we propose a novel evaluation framework for timeline construction to identify a better combination of a model and a formulation.
We further conduct benchmarking experiments on existing event temporal ordering datasets.
Based on our experiments focusing on open LLMs, we find that NLI formulation with Flan-T5 shows a strong performance among others, while timeline construction and event temporal ordering are both challenging tasks for few-shot LLMs. 
We believe that our work provides the research community with an evaluation testbed and insights into the current performance of LLMs for better timeline construction systems.

%% file: sections/limitation-and-ethical-consideration.tex
\section{Limitation}
Our TimeSET and all of the benchmarking datasets are based on English documents, specifically news articles.
Also, all the tasks are based on, at longest, a single document.
Our experimental results show that they are still challenging tasks for LLMs, even though the target setting (English, news, single document) is one of the popular domains in NLP.
Some previous studies expanded the frontiers: 
\citet{minard-etal-2016-meantime} includes multilingual documents;~\citet{minard-etal-2015-semeval} proposes a cross-document task;~\citet{bethard-etal-2012-annotating} targets novels.
Formulation comparison and benchmarking experiments on other settings would enhance the understanding of current LLMs' capabilities.

As for TimeSET, we manually choose documents, where we explicitly excluded some documents. 
For instance, we excluded obituaries and play-by-play sports reporting, in which textual order aligns almost perfectly with chronological order; and we excluded documents that describe complex parallel events, like epidemics where the jumble of events defy simple chronological order. 
In some cases for which we did not develop explicit criteria for exclusion, we may have implicitly favored specific types of documents where we easily spotted the chronology.
Hence, the timeline construction results could be different if formulation comparison experiments are conducted in more practical conditions, for instance, without any document restriction.  

Furthermore, our choice of LLMs (Llama 2 and Flan T5) is not exhaustive, although we aimed to cover model types as diverse as possible in an academic computational environment so that our findings can be generalized to other LLMs. 
As there exists a wide variety of promising both open and closed LLMs, we hope to include more diverse models in our future experiments.

For in-context learning, we randomly select demonstrations. However, demonstration selection strategies can have a large impact~\citep{min-etal-2022-rethinking}, and we leave also it to future work to examine the effect of the strategies on performance. 

Although this work focuses on event temporal relations, there are other types of event-to-event relations in event-centric IE, such as coreference relations and subevent relations. 
Future work can explore applying our framework of formulation comparison to such relations to obtain a further comprehensive understanding of LLMs on event-centric IE tasks. 

\section{Ethical Consideration}
In our formulation comparison and benchmarking experiments, we use (large) language models that are pretrained on a massive web-scraped corpus, which may contain some toxic or biased information. 
Although our target task, timeline construction, does not aim to generate any toxic outputs, some social/cultural biases in pretrained corpus could be cascaded to the outputs~\citep{gaut-etal-2020-towards}.

%% file: sections/appendix.tex
\section{TimeSET}
\label{appx:annotation}

\subsection{Additional Description about Annotation Guideline}
\label{appx:annotation_guideline}

\paragraph{Event}
When annotating salient events, annotators assume casual readers or people who do not have enough time to consume an entire content and think about what information should be shared with them. 
As for grammatical events,~\citet{minard-etal-2015-semeval} defines as follows: 
\begin{quote}
    Grammatical events are verbs or nouns that are semantically dependent on a governing content verb/noun. 
    Typical examples of grammatical events are copula verbs, light verbs followed by a nominal event, aspectual verbs and nouns, verbs and nouns expressing causal and motivational relations, and verbs and nouns expressing occurrence.
\end{quote}
We did not annotate types of events to reduce the annotation workload.
Among coreferring events, annotators can choose one based on the criteria of which mention is the most informative/representative. 
No annotation is added in title sentences.

\paragraph{Temporal link}
We introduce COEX link to represent a fuzzy relation where two events happened around the same time but their chronological relation is not mentioned explicitly in the context.
For instance, an article mentioned that both event X and event Y happened on one day, where readers can identify that the two events happened on the same day. 
However, the context may not tell the specific ordering of the events, possibly because the ordering is not important or unknown by the writer. 
In this case, we add a COEX link between event X and Y. 
To reduce annotation workload, we also introduce the concept of COEX clusters, the set of events connected with the COEX links.
As we can automatically expand AFTER links with the annotated COEX links, annotators need to annotate at least one preceding/following link for each cluster. 
Also, AFTER links can be annotated within one COEX cluster to enable accurate representation.

\paragraph{Argument}
Argument annotation contains two processes: 
(1) annotate an entity span.
(2) add an argument role link that connects an event with an entity. 
Entity spans can be multiple words.
Also, one entity span can be linked to multiple events with the same or different argument role links. 
Since we did not annotate an event type, argument roles can be inferred from multiple frames in Propbank. 
As role names, we use ``ARGX,'' where X can be 0 through 5, without descriptive names specified in each frame\footnote{For instance, in the frame of ``meet(.03)'', ``ARG0'' has additional descriptions of ``PAG'' (Prototypical AGent) and ``one party''}.

Note that we did not annotate time expressions in our dataset, as our focus is on reasonings based on context information without specific attention to time expressions. 

\input{tables/dataset_stats} 

\subsection{Annotation Interface}
\label{appx:annotation_interface}
\input{figures/annotation_interface}
We use two annotation interfaces, brat and our visualization tool. 
\autoref{fig:annotation_interface} shows the actual screenshot of each interface.
We customize the brat tool to support vertical lines for links.
Simultaneously, the visualization tool helps annotators verify their annotation in terms of how each event is connected to each other as a timeline.

\subsection{Document topics}
\label{appx:annotation_topics}
\input{tables/topics}

\autoref{table:topic} lists all the topics we cover in the TimeSET. 
In addition to the topic diversity, when available, we collected articles that described an event that happened in countries where English is not an official language.

\subsection{Detailed Statistics}
\label{appx:annotation_detailed_stats}

\autoref{table:dataset_stats} shows detailed statistics of TimeSET. 
We use \texttt{spaCy} to tokenize words. 

\subsection{Annotator}
Annotators are two experienced in-house knowledge engineers, paid adequately. 

\subsection{Annotation Analysis}
\label{appx:annotation_analysis}
In IAA calculation, we first resolve event coreference between annotations, considering our annotators choose the most representative mention among its coreference cluster.
This is motivated by the phenomenon that there can be multiple spans in one document that are as informative as each other (for instance, ``meeting'' and ``met.'').
Without accounting for this aspect, event annotations that refer to the same event but are labeled by different spans (i.e., text positions) get penalized.
As we do not annotate coreference links to keep annotation tractable, we manually identify coreferring events for IAA calculation.
3 documents in the test split are used for IAA calculations. 

As mentioned earlier, the pairwise F1 score for event spans is $0.74$ \footnote{$0.54$ without coreference resolution}.
Based on our manual analysis, the annotation discrepancy between annotators is largely attributed to the subjectivity in salient event selection. 
We addressed this point while developing our annotation guideline, however, we found it challenging to have a perfect match even with expert annotators.
This may imply that diversity inherently exists in the task of salience selection.

For temporal links, we use the temporal awareness score. 
There are a few variants of this score: 
The original version~\citep{uzzaman-allen-2011-temporal} uses predicted pairs as is.
The subsequent version~\citep{uzzaman-etal-2013-semeval} employs transitive reduction before calculating the score. 
The motivation of this idea is to appropriately reward explicit relation annotations more than implicit ones, which is suitable for the datasets where annotators choose explicitly which pair to annotate, like TimeBank.
However, in TimeSET, the implicit relations are as important as the explicit ones, so instead, we update the formula to use the transitive closures of both predicted and gold (annotated) graphs:
\begin{equation*}
    \text{Precision} = \frac{|G_1^+ \cap G_2^+|}{|G_1^+|}, \hspace{0.2cm}
    \text{Recall} = \frac{|G_2^+ \cap G_1^+|}{|G_2^+|}
\end{equation*}
\noindent where $G_1$ and $G_2$ are the timelines (graphs) from annotation compared to each other, and $G_i^+$ ($i=1,2$) is its transitive closure.
With this updated metric, we calculate IAA on temporal links for (1) all events, which contain the events one annotator added but not the other, and (2) only common events, where both of our expert annotators annotated as events.
Our motivation is to separate the effect of annotation discrepancy in event spans from temporal link annotation. 
The scores are $0.50$ and $0.90$, respectively \footnote{$0.30$ and $0.51$ without coreference resolution}.
This suggests that our temporal-link annotation itself is of high quality (i.e., IAA $=0.90$), which gets a propagated impact from the event span annotation (i.e., IAA $=0.50$).
We also note that in our formulation comparison experiment (Section~\ref{sec:formulation_comparison}), we give event spans as input together with context. 
This means that our experiment is mainly based on our high-quality (i.e., IAA $=0.90$) temporal link annotations.

\section{Formulation Comparison}
\label{appx:formulation_comparison}

\subsection{Experimental Details}
\label{appx:formulation_experiment_detail}

We load all models from~\url{https://huggingface.co/models} in brain floating point, \textit{bfloat16} and run inference or training. 
We use the \texttt{vllm} library for the inference of decoder-only models, and the \texttt{transformers} library for the inference of encoder-decoder models.
We set 4096 as the maximum input token length across all models. 
As described, the numbers of demonstrations we tested are 0, 1, and 2, which are relatively small numbers of demonstrations. 
We are constrained to use less than 3 demonstrations, as a document-level context in each demonstration fills up the input capacity.
To indicate target events, we add event indices surrounding each target event, for instance, ``[e1] investigate [/e1].'' 
contrary to~\citet{baldini-soares-etal-2019-matching}, we use ``['' instead of ``<'' because some of our target models, T5 and Flan-T5, do not have ``<'' in their vocabularies.
The indices are based on the textual order of events, starting from 1 to $n$, so that models cannot use the indices as cues in their inference. 
In pairwise, the indices are based on the order in each instance.  

As an evaluation metric, we use the pairwise F1 score based on all pairs averaged across documents.
This can be also seen as a variant of the temporal awareness score. 
One difference is that we do not calculate the transitive reduction from predicted timelines because the predicted timelines may have cycles.
One potential approach to remove potential cycles is to enforce consistency by using Integer Linear Programming~\citep{chambers-jurafsky-2008-jointly, han-etal-2019-joint}. 
In this work, we did not use ILP as our focus is to evaluate the bare capabilities of LLMs. 
We leave it to future work to assess the effect of the enforcement of consistency with LLMs.
For the Timeline formulation, we post-process generated timelines into a directed acyclic graph and we identify the temporal ordering of a pair based on node-to-node reachability. We use \texttt{networkx} library for graph-based processing.
For other formulations that may generate multiple relations for each pair, we take a majority vote for each, and if there is a tie, we select one randomly from the tied options.  

\subsection{Additional Analysis}
\label{appx:formulation_comparison_analysis}

\paragraph{Does the Number of Demonstrations Impact Performance?}
\input{figures/analysis_demonstration}
\autoref{fig:analysis_demo} shows the comparison among different numbers of demonstrations. 
We found that few-shot settings consistently outperform zero-shot settings.
However, increasing the number of demonstrations provides only marginal improvements or sometimes degrades performance. 

\paragraph{Does Event Representation Affect Performance?}
\input{figures/analysis_representation}
In our main experiments, we represent events with mentions (\textit{mention}) and event indices (\textit{eid}) like ``[e2]doping[/e2].'' 
We further investigate the effect of event representations on performance. 
As comparisons, we replace mention-based representation with a mention plus arguments (\textit{structured}), like ``[EVENT]doping[ARG1]Amir Khan[ARG2]...,'' inspired by~\citet{lin-etal-2021-conditional}. 
Also, we replace event indices with double asterisks (\textit{star}), inspired by the word-highlighting method in markdown files~\citep{wang-etal-2023-code4struct}. 
According to Figure~\ref{fig:analysis_repr}, we found that \textit{structured} representation or \textit{star} did not improve the performance.

\paragraph{Does Chain-of-thought or Code Prompt Improve Performance?}
\input{figures/analysis_prompt}
Previous studies reported the efficacy of adding chain-of-thoughts (CoT)~\citep{wei2022chain}.
We use CoT for MRC and timeline formulations, as they address multiple relations in one prediction. 
We transform the original pairwise annotations into sentences, like ``[e5]loss[/e5] started after [e2]doping[/e2]'' and use them as rationals (Table~\ref{table:prompt_template_mrc_cot}, \ref{table:prompt_template_timeline_cot})
Additionally, we tested code prompts, motivated by prior work, where code prompts improve performance in structured prediction tasks~\citep{madaan-etal-2022-language, wang-etal-2023-code4struct, zhang-etal-2023-causal}
We curated prompt templates, which have Python-like class structures, and applied them to the timeline formulation.
We use them for Timeline formulation, as it has the richest output structure (Table~\ref{table:prompt_template_timeline_code_template}).
As shown in Figure~\ref{fig:analysis_prompt}, prompts with CoT or code prompts perform on par with other prompts, even though CoT consumes additional input spaces.

\paragraph{Does the number of events affect performance?}
\input{figures/analysis_num_event}
Based on~\autoref{fig:analysis_num_event}, Performance gets worse with more events. Similar to the length of the document, this trend is more pronounced with formulations that identify multiple orders at once, like the Timeline formulation.

\section{Benchmarking}
\label{appx:benchmark}
Following are the additional details about our benchmarking experiments.

\subsection{Additional Description about Datasets}
\label{appx:benchmark_dataset}

\paragraph{MATRES} introduced multi-axis modeling, which maps events into the \textit{main}, \textit{intentional}, \textit{opinion}, and \textit{hypothetical} axis and annotates event pairs only within the same axis. 
Using crowd workers and focusing only on the event starting points, it became a large-scale dataset, achieving higher inter-annotator agreement than previous datasets, such as TempEval3~\citep{uzzaman-etal-2013-semeval}. 
As same as previous studies, they keep their context window as at most two adjacent sentences.

The task is, for a pair of events, to choose the temporal relationship from AFTER, BEFORE, EQUAL, and VAGUE, given a context and (the positions of) two events in the context. For instance,

\begin{quote}
    Context: Serbian police [e1]tried[/e1] to eliminate the pro-independence Kosovo Liberation Army and restore order. At least 51 people were [e2]killed[/e2] in clashes between Serb police and ethnic Albanians in the troubled region. \\
    Relation: BEFORE (e1 happened before e2)
\end{quote}

\noindent Following the previous studies~\citep{ning-etal-2019-improved, mathur-etal-2021-timers, zhou-etal-2022-rsgt}, we treat the ``AQUAINT'' and ``Timebank'' sets as our training split with 20\% of them as developing split, and use the ``Plutinum'' set as the test split. 
We measure performance with micro-F1 as same as those studies.
We use the annotation released in~\url{https://github.com/qiangning/MATRES}. 
No license is specified either in the paper or in the repository.
As the source documents are no longer available in the original place~\url{https://www.cs.york.ac.uk/semeval-2013/task1/}, we download the copy from~\url{https://github.com/qiangning/StructTempRel-EMNLP17/tree/master/data/TempEval3}.
We couldn't find the license of the source documents (TempEval 3) either in their paper. 

\paragraph{TDDiscourse} introduced discourse-level pairwise event temporal relation annotation to capture long-context temporal relations on top of TBDense~\citep{cassidy-etal-2014-annotation}. 
Specifically, their annotation contains temporal relations between event pairs, where they are more than one sentence apart. 
It has five temporal relations, BEFORE, AFTER, SIMULTANEOUS, INCLUDE, and INCLUDED. 
We use the manually-annotated set, TDDMan, in our experiment, while there is another set, TDDAuto, which is an automatically-induced annotation based on temporal expressions~\citep{reimers-etal-2016-temporal}. 
This decision is based on our aim of focusing on event temporal ordering in general, not specific to the relations that can be inferred from time expressions. 
Additionally, the authors reported that TDDMan contains more difficult phenomena than TDDAuto, which makes it more suitable as a testbed. 

Similar to the MATRES dataset, the task is to identify the temporal relationship between a pair of events, based on its context. For instance, 

\begin{quote}
    Context(snippet): Atlanta nineteen ninety-six. A bomb [e1]blast[/e1] shocks the Olympic games. One person is killed. January nineteen ninety-seven. Atlanta again. This time a bomb at an abortion clinic. More people are [e2]hurt[/e2]. \\
    Relation: BEFORE (e1 happened before e2)
\end{quote}

We use the original split of train, development, and test split.
The annotation is downloaded from~\url{https://github.com/aakanksha19/TDDiscourse} and the source data is from the same place as MATRES. No license information is specified in either the paper or the repository.

\paragraph{TemporalNLI} recast four event temporal ordering datasets into NLI formats to better understand the temporal reasoning capabilities of  NLI models trained on common NLI datasets. 
Original annotations are converted automatically in a rule-based manner, where they used eight hypothesis templates, such as ``X started before Y ended'' or ``Y ended before X started.''

The task is to output a binary label, entailment or not entailment, given a context and a hypothesis, for instance, 

\begin{quote}
    Context: Reggie said he will pay us soon. \\
    Hypothesis: The paying ended before the saying started. \\
    Label: Not entailment
\end{quote}

Among the subsets, we use the recast version of TempEval3 to assess models consistently on similar documents (MATRES is a refined version of TempEval3) and TempEval3 contains TBDense documents. 
We follow the code and instructions available in~\url{https://github.com/sidsvash26/temporal_nli} (MIT License) to obtain the data.
We use the provided data split in our experiment.
The source documents are retrieved in the same way as in MATRES.

\paragraph{TORQUE} introduced a machine reading comprehension dataset to test temporal relation understanding. 
This machine reading comprehension formulation gives two unique characteristics, compared to other existing pairwise relation classification datasets: 
(1) The formulation can query multiple relations at one time, i.e., one-to-many relationships between the event in a question and events in a context. 
(2) A natural language question can express fuzzy relations, e.g., ``What may have happened when [event] happened?'' (uncertain) or ``What did not start after [event]?'' (negation). 
The contexts are samples from the source documents of TempEval3.

The task is to extract event spans from a context, given the context and one temporal reasoning question, for instance:

\begin{quote}
    Context: Heavy snow is causing disruption to transport across the UK, with heavy rainfall bringing flooding to the south-west of England. Rescuers searching for a woman trapped in a landslide at her home in Looe, Cornwall, said they had found a body. \\
    Question: What happened after a woman was trapped? \\
    Answer:  searching, said, found
\end{quote}

We download the annotated data available in~\url{https://github.com/qiangning/TORQUE-dataset} (Apache License 2.0) and use the original data split. 
As annotation on the test split is not publicly available (only available via leaderboard submission~\url{https://leaderboard.allenai.org/torque/submissions/public}), we further split the train set into our original training set (80\%) and validation set (20\%) and report performance on the original development set.
We use the original annotation as is, contrary to the one after the rule-based automatic conversion~\url{https://github.com/rujunhan/TORQUE} to avoid introducing any biases from the post-process. 
In our evaluation, we use the same strategy as the previous works~\citep{ning-etal-2020-torque, huang-etal-2022-understand}: one prediction obtained from each instance is evaluated against all individual annotations and the best score is used for overall results. 

Other than event-centric IE datasets, BBH~\citep{suzgun-etal-2023-challenging} contains not IE-specific temporal reasoning tasks, such as the \textit{Data Understanding} task and \textit{Temporal Sequences} task. 
Our dataset with a specific focus on events and IE can be complementary to such tasks. 
Also, it would be beneficial to expand benchmarking experiments with other event temporal ordering datasets using one standard split~\citep{sousa2023tieval} for further comprehensive understanding.

\subsection{Finetuned Model Architecture}
\label{appx:benchmark_model_arch}
As for the encoder-only models, we re-implement a baseline system for each task.
We use the DeBERTaV3's Transformer layers as a base. 
For the pairwise relation extraction datasets (MATRES and TDDiscourse) we add a linear layer that projects the embeddings of the target event tokens to output labels (multi-class classification).
For the NLI dataset (TemporalNLI), we add a linear layer that projects sentence embedding to binary labels (binary classification task).
For the MRC dataset (TORQUE), we add a linear layer that classifies each token into a binary class, event or not event (sequence tagging task). 
Our motivation for using DeBERTaV3 over other competitive models, such as RoBERTa~\citep{liu2020roberta} or ELECTRA~\citep{clark2020electra}, is based on DeBERTaV3's support for longer input length.

\subsection{Experimental Details}
\label{appx:benchmark_experiment_detail}
\input{tables/hyperparameters}
We load all models from~\url{https://huggingface.co/models} in brain floating point, \textit{bfloat16} and run inference or training. 
We use the \texttt{vllm} library for the inference of decoder-only models, and the \texttt{transformers} library for the inference of other models and training models.
Contrary to the formulation comparison experiment, we did not include CodeLlama, considering that the target datasets in this experiment do not possess enough structures in output.
For in-context learning approaches, we use different numbers of A6000 GPUs, depending on the size of a model. 
The number of demonstrations are 0, 1, 3, 5, or 10.
For parameter-updating approaches, we use a single A6000 GPU for each training. 
Computational time varies depending on model architecture, model size, and dataset. On average, one configuration takes around a few hours. 
We conduct a grid search for the following hyper-parameters (numbers within parentheses are their options): learning rate (1e-4, 1e-5, and 1e-6), and batch size (8, 16, and 32). 
Additionally, for LoRA, we tune dimensions (16 and 64) and alphas (16 and 64).
We set the maximum number of epochs as 10, and apply early stopping. 
Hyperparameters for the reported results are available in~\autoref{table:hyperparameters}.
We use string match as employed in HELM~\citep{liang2022holistic} over likelihood-based evaluation~\citep{hendrycks2021measuring, eval-harness} because likelihood-based (ranking-based) evaluation is not always a straightforward approach, for instance, in MRC settings, where the search space for answers can be rather large. 
Also, although the focus of this work is open models, likelihoods may not be always available from API calls when evaluating closed models.
Hence, we use string-based evaluation as a less-constraint approach. 
As evaluation approaches can have non-negligible effects on performance~\footnote{https://huggingface.co/blog/evaluating-mmlu-leaderboard}, we leave it as a future work to investigate how much effect evaluation strategy has in event temporal ordering tasks. 

\section{Prompt Templates}
\label{appx:prompt_template}
One prompt template and example demonstration for each formulation and prompt type are shown in Table~\ref{table:prompt_template_nli} through~\ref{table:prompt_template_timeline_code_demonstration}.

\input{tables/prompt_template_nli}

\input{tables/prompt_template_pairwise}

\input{tables/prompt_template_mrc}

\input{tables/prompt_template_mrc_cot}

\input{tables/prompt_template_timeline}

\input{tables/prompt_template_timeline_cot}

\input{tables/prompt_template_timeline_code}

%% file: tables/dataset_stats.tex
\begin{table}[!t]
\centering
\begin{adjustbox}{width=0.5\textwidth}
\begin{tabular}{l c c c}
    \toprule
    & dev & test & all \\
    \midrule
    \#docs & 10 & 40 & 50 \\
    \#word & (458.4) & (432.8) & (437.9) \\
    \#sentence & (16.8) & (15.0) & (15.3) \\
    \midrule
    \#event & 91 (9.1) & 265 (6.6) & 356 (7.1)\\
    \#relation & 81 (8.1) & 233 (5.8) & 314 (6.3) \\
    \#argument & 166 (16.6) & 488 (12.2) & 654 (13.1) \\
    \bottomrule
\end{tabular}
\end{adjustbox}
\caption{Dataset Statistics. Numbers in parentheses are the averages per document.}
\label{table:dataset_stats}
\end{table}

%% file: figures/annotation_interface.tex
\begin{figure*}[t]
    \centering
    \begin{subfigure}[b]{\textwidth}
        \centering
        \includegraphics[width=\textwidth]{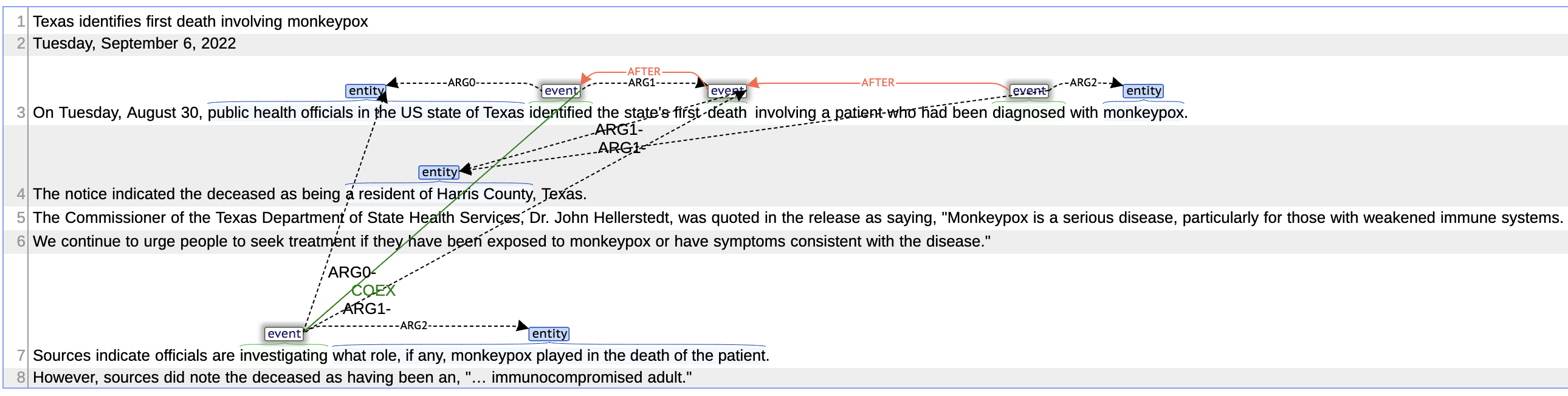}
        \caption{Brat Inferface}
        \label{fig:brat}
    \end{subfigure}    
    
    \vspace{0.5cm}
    
    \begin{subfigure}[b]{0.5\textwidth}
        \centering
        \includegraphics[width=\textwidth]{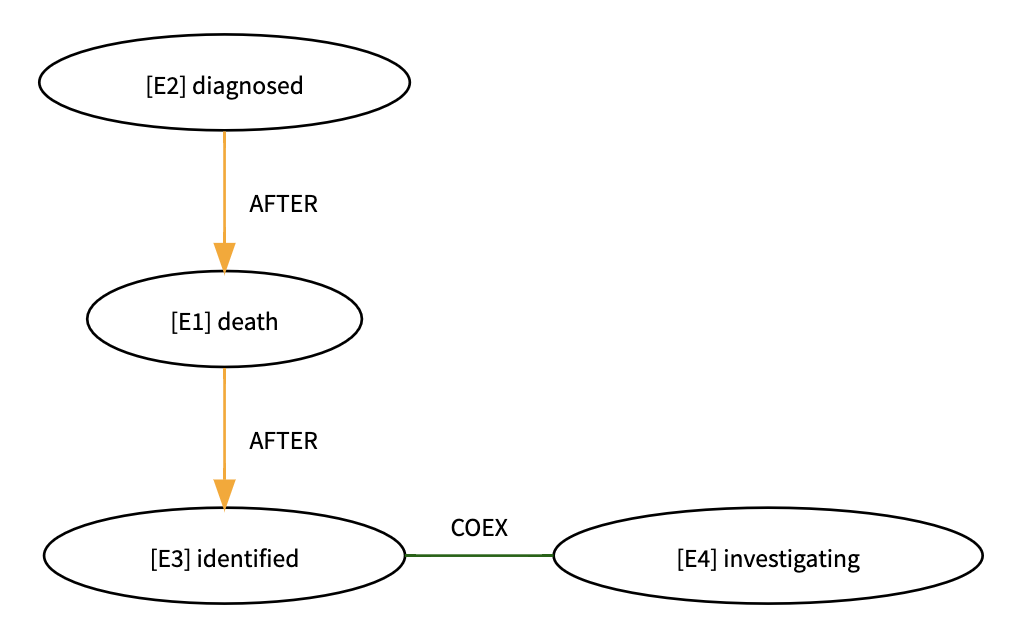}
        \caption{Visualization Interface}
        \label{fig:vis}
    \end{subfigure}
    \caption{Annotation Interface}
    \label{fig:annotation_interface}
\end{figure*}

%% file: tables/topics.tex
\begin{table}[!t]
\centering
\begin{adjustbox}{width=0.45\textwidth}
\begin{tabular}{l}
    \toprule
     computers \\
     crime and law - accuse\\
     culture and entertainment\\
     disasters and accidents - road crash \\
     disasters and accidents - shipwreck \\
     disasters and accidents - volcanoes \\
     economy and business - financial crisis \\
     economy and business - strikes \\
    environment \\
    government and politics - protests \\
    government and politics - riots
    health \\
    infectious disease \\
    internet \\
    mining \\
    politics and conflicts \\
    politics and conflicts - armed conflict \\
    politics and conflicts - drones \\
    politics and conflicts - elections \\
    politics and conflicts - policy change \\
    politics and conflicts - resign \\
    rail transport \\
    space \\
    sports \\
    weather - earthquakes \\
    weather - floods \\
    weather - storms \\
    \bottomrule
\end{tabular}
\end{adjustbox}
\caption{Topics}
\label{table:topic}
\end{table}

%% file: figures/analysis_demonstration.tex
\begin{figure}[t]
    \centering
    \includegraphics[width=0.5\textwidth]{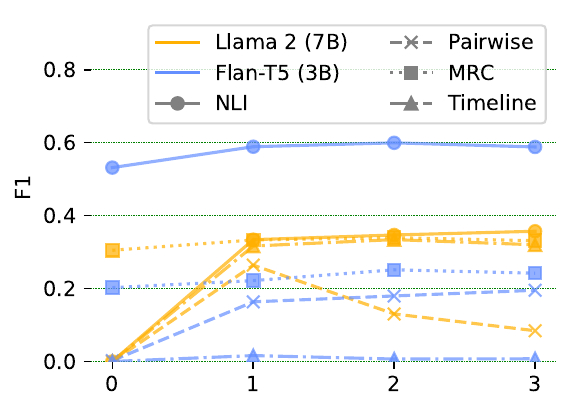}
    \caption{\#demonstration}
    \label{fig:analysis_demo}
\end{figure}

%% file: figures/analysis_representation.tex
\begin{figure}[t]
    \centering
    \includegraphics[width=0.5\textwidth]{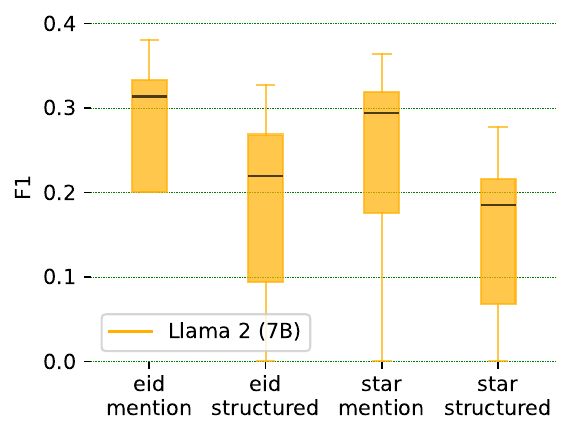}
    \caption{Event representation}
    \label{fig:analysis_repr}
\end{figure}

%% file: figures/analysis_prompt.tex
\begin{figure}[t]
    \centering
    \includegraphics[width=0.5\textwidth]{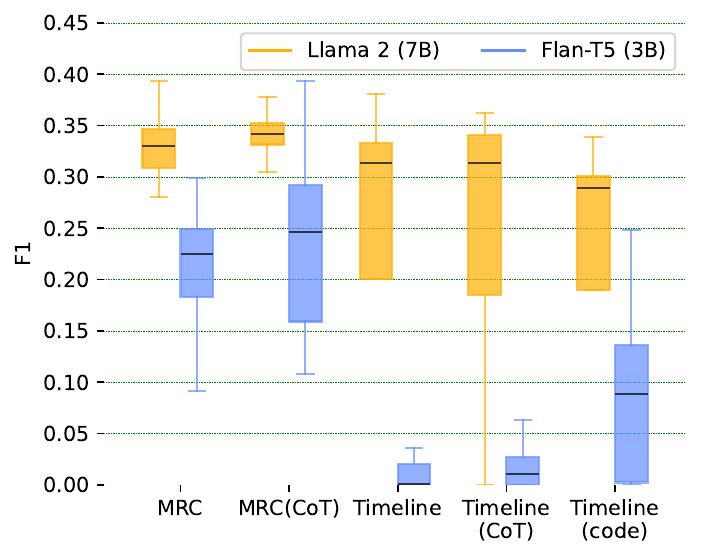}
    \caption{CoT and code prompt}
    \label{fig:analysis_prompt}
\end{figure}

%% file: figures/analysis_num_event.tex
\begin{figure}[t]
    \centering
    \includegraphics[width=0.5\textwidth]{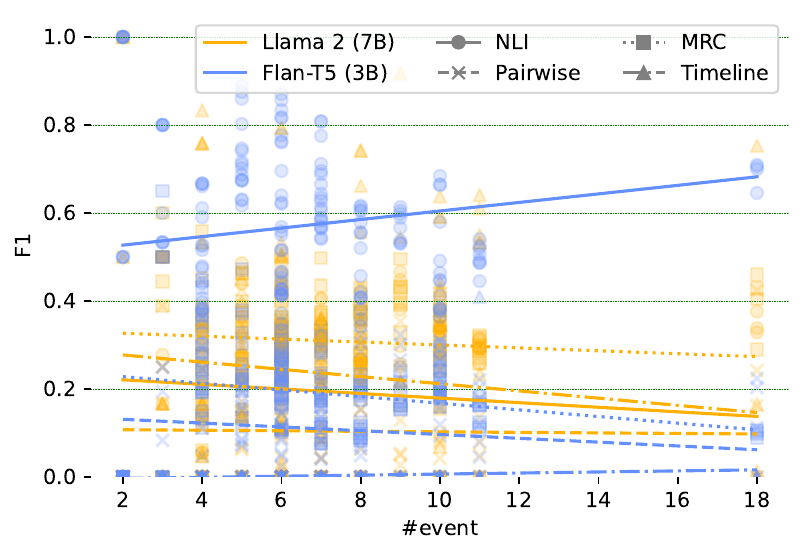}
    \caption{\#demonstration}
    \label{fig:analysis_num_event}
\end{figure}

%% file: tables/hyperparameters.tex
\begin{table*}[!t]
\centering
\begin{adjustbox}{width=0.9\textwidth}
\begin{tabular}{l c c c c c c}
    \toprule
    Dataset & Tuning Method & Model & Learning Rate & Batch Size & \makecell[c]{Dimentions\\(LoRA)} & \makecell[c]{Alpha\\(LoRA)} \\
    \midrule
    \multirow{6}{*}{TemporalNLI} & \multirow{2}{*}{FT} & DeBERTaV3 (440M) & 1e-04 & 16 & --- & --- \\
    & & Flan-T5 (770M) & 1e-04 & 16 & --- & --- \\
    \cmidrule(l){2-7}
    & \multirow{4}{*}{PEFT} & DeBERTaV3 (440M) & 1e-05 & 8 & 16 & 64 \\
    & & Flan-T5 (770M) & 1e-04 & 8 & 16 & 16 \\
    & & Flan-T5 (3B) & 1e-04 & 8 & 64 & 64 \\
    & & Llama-2 (7B) & 1e-04 & 8 & 64 & 16 \\
    \midrule
    \multirow{6}{*}{MATRES} & \multirow{2}{*}{FT} & DeBERTaV3 (440M) & 1e-04 & 32 & --- & --- \\
    & & Flan-T5 (770M) & 1e-04 & 16 & --- & --- \\
    \cmidrule(l){2-7}
    & \multirow{4}{*}{PEFT} & DeBERTaV3 (440M) & 1e-04 & 8 & 64 & 64 \\
    & & Flan-T5 (770M) & 1e-04 & 8 & 16 & 64 \\
    & & Flan-T5 (3B) & 1e-04 & 8 & 64 & 64 \\
    & & Llama-2 (7B) & 1e-04 & 8 & 16 & 64 \\
    \midrule
    \multirow{6}{*}{TDDiscourse} & \multirow{2}{*}{FT} & DeBERTaV3 (440M) & 1e-04 & 16 & --- & --- \\
    & & Flan-T5 (770M) & 1e-04 & 16 & --- & --- \\
    \cmidrule(l){2-7}
    & \multirow{4}{*}{PEFT} & DeBERTaV3 (440M) & 1e-04 & 8 & 64 & 16 \\
    & & Flan-T5 (770M) & 1e-04 & 8 & 64 & 16 \\
    & & Flan-T5 (3B) & 1e-04 & 8 & 64 & 64 \\
    & & Llama-2 (7B) & --- & --- & --- & --- \\
    \midrule
    \multirow{6}{*}{TORQUE} & \multirow{2}{*}{FT} & DeBERTaV3 (440M) & 1e-04 & 32 & --- & --- \\
    & & Flan-T5 (770M) & 1e-04 & 16 & --- & --- \\
    \cmidrule(l){2-7}
    & \multirow{4}{*}{PEFT} & DeBERTaV3 (440M) & 1e-04 & 8 & 16 & 64 \\
    & & Flan-T5 (770M) & 1e-04 & 8 & 64 & 64 \\
    & & Flan-T5 (3B) & 1e-04 & 8 & 64 & 64 \\
    & & Llama-2 (7B) & 1e-04 & 8 & 16 & 64 \\
    \bottomrule
\end{tabular}
\end{adjustbox}
\caption{Best hyperparameters for (parameter-efficient) fine-tuned models.}
\label{table:hyperparameters}
\end{table*}

%% file: tables/prompt_template_nli.tex
\begin{table*}[!t]
\centering
\begin{adjustbox}{width=\textwidth}
\begin{tabular}{p{15cm}}
    \toprule
    \makecell[c]{[Template]} \\
    We say that one sentence ``entails'' another sentence when the first sentence implies the second sentence. Consider the following two sentences: \\
    \{premise\} \\
    \{hypothesis\} \\
    Is the relationship from the first to the second sentence ``entailment'' or ``not entailment''? \\
    \{target\}\\
    \midrule
    \makecell[c]{[Example Demonstration]} \\
    We say that one sentence "entails" another sentence when the first sentence implies the second sentence. Consider the following two sentences:\\
    Texas identifies first death involving monkeypox\\
    Tuesday, September 6, 2022 \\
    On Tuesday, August 30, public health officials in the US state of Texas identified the state's first [e1]death[\/e1] involving a patient who had been diagnosed with monkeypox. The notice indicated the deceased as being a resident of Harris County, Texas.\\
    The Commissioner of the Texas Department of State Health Services, Dr. John Hellerstedt, was quoted in the release as saying, ``Monkeypox is a serious disease, particularly for those with weakened immune systems. We continue to urge people to seek treatment if they have been exposed to monkeypox or have symptoms consistent with the disease.''\\
    Sources indicate officials are [e2]investigating[\/e2] what role, if any, monkeypox played in the death of the patient. However, sources did note the deceased as having been an, ``... immunocompromised adult.''\\
    
    [e2]investigating[\/e2] started after [e1]death[\/e1] started.\\
    Is the relationship from the first to the second sentence ``entailment'' or ``not entailment''?\\
    Entailment\\
    \bottomrule
\end{tabular}
\end{adjustbox}
\caption{One prompt template and example demonstration: NLI}
\label{table:prompt_template_nli}
\end{table*}

%% file: tables/prompt_template_pairwise.tex
\begin{table*}[!t]
\centering
\begin{adjustbox}{width=\textwidth}
\begin{tabular}{p{15cm}}
    \toprule
    \makecell[c]{[Template]} \\
    Passage:\\
    \{context\}\\
    Based on the passage, choose the temporal relation of the two events, \{arg1\} and \{arg2\}?\\
    Please answer with one of the following options:\\
    \{choices\}\\
    Answer: \{target\}\\
    \midrule
    \makecell[c]{[Example Demonstration]} \\
    Texas identifies first death involving monkeypox\\
    Tuesday, September 6, 2022 \\
    On Tuesday, August 30, public health officials in the US state of Texas identified the state's first [e1]death[\/e1] involving a patient who had been diagnosed with monkeypox. The notice indicated the deceased as being a resident of Harris County, Texas.\\
    The Commissioner of the Texas Department of State Health Services, Dr. John Hellerstedt, was quoted in the release as saying, "Monkeypox is a serious disease, particularly for those with weakened immune systems. We continue to urge people to seek treatment if they have been exposed to monkeypox or have symptoms consistent with the disease."\\
    Sources indicate officials are [e2]investigating[\/e2] what role, if any, monkeypox played in the death of the patient. However, sources did note the deceased as having been an, "… immunocompromised adult."\\
    Based on the passage, choose the temporal relation of the two events, [e2]investigating[\/e2] and [e1]death[\/e1]?\\
    Please answer with one of the following options:\\
    - BEFORE: The first event started before the second event started.\\
    - AFTER: The first event started after the second event started.\\
    - VAGUE: The temporal relationship between the first and second event is unclear.\\
    Answer: AFTER\\
    \bottomrule
\end{tabular}
\end{adjustbox}
\caption{One prompt template and example demonstration: Pairwise}
\label{table:prompt_template_pairwise}
\end{table*}

%% file: tables/prompt_template_mrc.tex
\begin{table*}[!t]
\centering
\begin{adjustbox}{width=\textwidth}
\begin{tabular}{p{15cm}}
    \toprule
    \makecell[c]{[Template]} \\
    Answer the following question, ``\{question\}'' using the information below.\\
    \{context\}\\
    Hint: events are marked with ``\{marker\}''.\\
    Answer: \{target\}\\
    \midrule
    \makecell[c]{[Example Demonstration]} \\
    Answer the following question, ``Which events happened before [e1]identified[\/e1]?'' using the information below.\\
    Texas identifies first death involving monkeypox\\
    Tuesday, September 6, 2022 \\
    On Tuesday, August 30, public health officials in the US state of Texas [e1]identified[\/e1] the state's first [e2]death[\/e2] involving a patient who had been [e3]diagnosed[\/e3] with monkeypox. The notice indicated the deceased as being a resident of Harris County, Texas.\\
    The Commissioner of the Texas Department of State Health Services, Dr. John Hellerstedt, was quoted in the release as saying, ``Monkeypox is a serious disease, particularly for those with weakened immune systems. We continue to urge people to seek treatment if they have been exposed to monkeypox or have symptoms consistent with the disease.''\\
    Sources indicate officials are [e4]investigating[\/e4] what role, if any, monkeypox played in the death of the patient. However, sources did note the deceased as having been an, "… immunocompromised adult."\\
    Hint: events are marked with ``[e]''.\\
    Answer: [e2]death[\/e2], [e3]diagnosed[\/e3]\\
    \bottomrule
\end{tabular}
\end{adjustbox}
\caption{One prompt template and example demonstration: MRC}
\label{table:prompt_template_mrc}
\end{table*}

%% file: tables/prompt_template_mrc_cot.tex
\begin{table*}[!t]
\centering
\begin{adjustbox}{width=\textwidth}
\begin{tabular}{p{15cm}}
    \toprule
    \makecell[c]{[Template]} \\
    Context:\\
    \{context\}\\
    Question: \{question\}\\
    Answer candidates:\\
    \{events\}\\
    Chain of thoughts:\\
    \{cot\}\\
    Answer:\\
    \{target\}\\
    \midrule
    \makecell[c]{[Example Demonstration]} \\
    Context:\\
    Texas identifies first death involving monkeypox\\
    Tuesday, September 6, 2022 \\
    On Tuesday, August 30, public health officials in the US state of Texas [e1]identified[\/e1] the state's first [e2]death[\/e2] involving a patient who had been [e3]diagnosed[\/e3] with monkeypox. The notice indicated the deceased as being a resident of Harris County, Texas.\\
    The Commissioner of the Texas Department of State Health Services, Dr. John Hellerstedt, was quoted in the release as saying, ``Monkeypox is a serious disease, particularly for those with weakened immune systems. We continue to urge people to seek treatment if they have been exposed to monkeypox or have symptoms consistent with the disease.''\\
    Sources indicate officials are [e4]investigating[\/e4] what role, if any, monkeypox played in the death of the patient. However, sources did note the deceased as having been an, ``… immunocompromised adult.''\\
    Question: Which events happened before [e1]identified[\/e1]?\\
    Answer candidates:\\
    - [e1]identified[\/e1]\\
    - [e2]death[\/e2]\\
    - [e3]diagnosed[\/e3]\\
    - [e4]investigating[\/e4]\\
    Chain of thoughts: \\
    
    [e2]death[\/e2] happened before [e1]identified[\/e1]. [e3]diagnosed[\/e3] happened before [e2]death[\/e2]. [e4]investigating[\/e4] does not have a clear temporal relation with [e1]identified[\/e1]. \\
    Answer:\\
    - [e2]death[\/e2], [e3]diagnosed[\/e3]\\
    \bottomrule
\end{tabular}
\end{adjustbox}
\caption{One prompt template and example demonstration: MRC (CoT)}
\label{table:prompt_template_mrc_cot}
\end{table*}

%% file: tables/prompt_template_timeline.tex
\begin{table*}[!t]
\centering
\begin{adjustbox}{width=\textwidth}
\begin{tabular}{p{15cm}}
    \toprule
    \makecell[c]{[Template]} \\
    Please create a timeline based on a context. A timeline is a linear representation of events, ordered chronologically. You can put multiple events in one time span if they happened around the same time.\\
    Given the passage below:\\
    \{context\}\\
    You can identify the following events:\\
    \{events\}\\
    Now, create a timeline with the events:\\
    \{target\}\\
    \midrule
    \makecell[c]{[Example Demonstration]} \\
    Please create a timeline based on a context. A timeline is a linear representation of events, ordered chronologically. You can put multiple events in one time span if they happened around the same time.\\
    Given the passage below:\\
    Texas identifies first death involving monkeypox\\
    Tuesday, September 6, 2022 \\
    On Tuesday, August 30, public health officials in the US state of Texas [e1]identified[\/e1] the state's first [e2]death[\/e2] involving a patient who had been [e3]diagnosed[\/e3] with monkeypox. The notice indicated the deceased as being a resident of Harris County, Texas.\\
    The Commissioner of the Texas Department of State Health Services, Dr. John Hellerstedt, was quoted in the release as saying, ``Monkeypox is a serious disease, particularly for those with weakened immune systems. We continue to urge people to seek treatment if they have been exposed to monkeypox or have symptoms consistent with the disease.''\\
    Sources indicate officials are [e4]investigating[\/e4] what role, if any, monkeypox played in the death of the patient. However, sources did note the deceased as having been an, ``… immunocompromised adult.''\\
    You can identify the following events:\\
    - [e1]identified[\/e1]\\
    - [e2]death[\/e2]\\
    - [e3]diagnosed[\/e3]\\
    - [e4]investigating[\/e4]\\
    Now, create a timeline with the events:\\
    T1:\\
    - [e3]diagnosed[\/e3]\\
    T2:\\
    - [e2]death[\/e2]\\
    T3:\\
    - [e1]identified[\/e1]\\
    - [e4]investigating[\/e4]\\
    \bottomrule
\end{tabular}
\end{adjustbox}
\caption{One prompt template and example demonstration: Timeline}
\label{table:prompt_template_timeline}
\end{table*}

%% file: tables/prompt_template_timeline_cot.tex
\begin{table*}[!t]
\centering
\begin{adjustbox}{width=\textwidth}
\begin{tabular}{p{15cm}}
    \toprule
    \makecell[c]{[Template]} \\
    A timeline consists of multiple time spans, which ordered chronologically. Each time span contains one or more events that happened around the same time without clear temporal relations. Your task is to create a timeline from the next article.\\
    Article:\\
    \{context\}\\
    Suppose you identified the following events from the article above:\\
    \{events\}\\
    Additionally, you identified the following temporal relations in the article:\\
    \{cot\}\\
    Then, the timeline is:\\
    \{target\}\\
    \midrule
    \makecell[c]{[Example Demonstration]} \\
    A timeline consists of multiple time spans, which ordered chronologically. Each time span contains one or more events that happened around the same time without clear temporal relations. Your task is to create a timeline from the next article.\\
    Article:\\
    Texas identifies first death involving monkeypox\\
    Tuesday, September 6, 2022 \\
    On Tuesday, August 30, public health officials in the US state of Texas [e1]identified[\/e1] the state's first [e2]death[\/e2] involving a patient who had been [e3]diagnosed[\/e3] with monkeypox. The notice indicated the deceased as being a resident of Harris County, Texas.\\
    The Commissioner of the Texas Department of State Health Services, Dr. John Hellerstedt, was quoted in the release as saying, "Monkeypox is a serious disease, particularly for those with weakened immune systems. We continue to urge people to seek treatment if they have been exposed to monkeypox or have symptoms consistent with the disease."\\
    Sources indicate officials are [e4]investigating[\/e4] what role, if any, monkeypox played in the death of the patient. However, sources did note the deceased as having been an, "… immunocompromised adult."\\
    Suppose you identified the following events from the article above:\\
    - [e1]identified[\/e1]\\
    - [e2]death[\/e2]\\
    - [e3]diagnosed[\/e3]\\
    - [e4]investigating[\/e4]\\
    Additionally, you identified the following temporal relations in the article:\\
    
    [e2]death[\/e2] started after [e3]diagnosed[\/e3]. [e1]identified[\/e1] started after [e2]death[\/e2]. [e1]identified[\/e1] started around the same time, but the temporal relationship with [e4]investigating[\/e4] is not clear. \\
    Then, the timeline is:\\
    T1:\\
    - [e3]diagnosed[\/e3]\\
    T2:\\
    - [e2]death[\/e2]\\
    T3:\\
    - [e1]identified[\/e1]\\
    - [e4]investigating[\/e4]\\
    \bottomrule
\end{tabular}
\end{adjustbox}
\caption{One prompt template and example demonstration: Timeline (CoT)}
\label{table:prompt_template_timeline_cot}
\end{table*}

%% file: tables/prompt_template_timeline_code.tex
\begin{table*}[!t]
\centering
\begin{adjustbox}{width=\textwidth}
\begin{tabular}{p{15cm}}
    \toprule
    \makecell[c]{[Template]} \\
    """\\
    Context:\\
    \{context\}\\
    """\\
    class Timeline:\\
    \hspace{0.5cm}    """\\
    \hspace{0.5cm}    This class is the structured representation of the context above.\\
    \hspace{0.5cm}    * Events are the actions or happenings in it.\\
    \hspace{0.5cm}    * Timeline is a chronologically ordered events, where two events that happened around the same time are placed in the same time span.\\
    \hspace{0.5cm}        A time span is represented by T[num].\\
    \\
    \hspace{0.5cm}     """\\
    \hspace{0.5cm}     def \_\_init\_\_(self):\\
    \hspace{1cm}         \# Events\\
    \{events\}\\
    \\
    \hspace{1cm}         \# Timeline\\
    \{target\}\\
    \midrule
    \bottomrule
\end{tabular}
\end{adjustbox}
\caption{One prompt template: Timeline (code)}
\label{table:prompt_template_timeline_code_template}
\end{table*}

\begin{table*}[!t]
\centering
\begin{adjustbox}{width=\textwidth}
\begin{tabular}{p{15cm}}
    \toprule
    \makecell[c]{[Example Demonstration]} \\
    """\\
    Context:\\
    Texas identifies first death involving monkeypox\\
    Tuesday, September 6, 2022 \\
    On Tuesday, August 30, public health officials in the US state of Texas [e1]identified[\/e1] the state's first [e2]death[\/e2] involving a patient who had been [e3]diagnosed[\/e3] with monkeypox. The notice indicated the deceased as being a resident of Harris County, Texas.\\
    The Commissioner of the Texas Department of State Health Services, Dr. John Hellerstedt, was quoted in the release as saying, "Monkeypox is a serious disease, particularly for those with weakened immune systems. We continue to urge people to seek treatment if they have been exposed to monkeypox or have symptoms consistent with the disease."\\
    Sources indicate officials are [e4]investigating[\/e4] what role, if any, monkeypox played in the death of the patient. However, sources did note the deceased as having been an, "… immunocompromised adult."\\
    """\\
    class Timeline:\\
    \hspace{0.5cm}    """\\
    \hspace{0.5cm}    This class is the structured representation of the context above.\\
    \hspace{0.5cm}    * Events are the actions or happenings in it.\\
    \hspace{0.5cm}    * Timeline is a chronologically ordered events, where two events that happened around the same time are placed in the same time span.\\
    \hspace{0.5cm}        A time span is represented by T[num].\\
    \\
   \hspace{0.5cm}     """\\
    \hspace{0.5cm}    def \_\_init\_\_(self):\\
    \hspace{1cm}        \# Events\\
    \hspace{1cm}        self.event1 = events[1] \# [e1]identified[\/e1]\\
    \hspace{1cm}        self.event2 = events[2] \# [e2]death[\/e2]\\
    \hspace{1cm}        self.event3 = events[3] \# [e3]diagnosed[\/e3]\\
    \hspace{1cm}        self.event4 = events[4] \# [e4]investigating[\/e4]\\
    \\
    \hspace{1cm}        \# Timeline\\
    \hspace{1cm}        T1 = [self.event3]\\
    \hspace{1cm}        T2 = [self.event2]\\
    \hspace{1cm}        T3 = [self.event1, self.event4]\\
    \bottomrule
\end{tabular}
\end{adjustbox}
\caption{One example demonstration: Timeline (code)}
\label{table:prompt_template_timeline_code_demonstration}
\end{table*}